\documentclass[lettersize,journal]{IEEEtran}
\usepackage{amsmath,amsfonts}
\usepackage{algorithmic}
\usepackage{array}
\usepackage[caption=false,font=normalsize,labelfont=sf,textfont=sf]{subfig}
\usepackage{textcomp}
\usepackage{stfloats}
\usepackage{url}
\usepackage{verbatim}
\usepackage{graphicx}

\usepackage{multirow}
\usepackage{amsfonts} 

\def\BibTeX{{\rm B\kern-.05em{\sc i\kern-.025em b}\kern-.08em
    T\kern-.1667em\lower.7ex\hbox{E}\kern-.125emX}}
\usepackage{balance}
\begin{document}
\title{6-DoF Robotic Grasping with Transformer}
\author{Zhenjie Zhao, Hang Yu, Hang Wu$^{*}$, Xuebo Zhang$^{*}$
\thanks{
${*}$: corresponding authors

Zhenjie Zhao, Hang Yu, Hang Wu, and Xuebo Zhang are with the
College of Artificial Intelligence, Institute of Robotics and Automatic Information System, and the Tianjin Key Laboratory of Intelligent Robotics, Nankai
University, Tianjin 300350, China (e-mail: zhangxuebo@nankai.edu.cn).}}


\maketitle

\begin{abstract}
Robotic grasping aims to detect graspable points and their corresponding gripper configurations in a particular scene, 
and is fundamental for robot manipulation. Existing research works have demonstrated the potential of using 
a transformer model for robotic grasping, which can efficiently learn both global and local features.
However, such methods are still limited in grasp detection on a 2D plane. 
In this paper, we extend a transformer model for 6-Degree-of-Freedom (6-DoF) robotic grasping, 
which makes it more flexible and suitable for tasks that concern safety. 
The key designs of our method are a serialization module that turns a 3D voxelized space into a sequence of feature tokens 
that a transformer model can consume
and skip-connections 
that merge multiscale features effectively.
In particular, our method takes a Truncated Signed Distance Function (TSDF) as input.
After serializing the TSDF, a transformer model is utilized to encode the sequence,
which can obtain a set of aggregated hidden feature vectors through multi-head attention.
We then decode the hidden features to obtain per-voxel feature vectors through deconvolution and skip-connections. 
Voxel feature vectors are then used to regress parameters for executing grasping actions.
On a recently proposed pile and packed grasping dataset, we showcase that our transformer-based method 
can surpass existing methods by about 5\% in terms of success rates and declutter rates. 
We further evaluate the running time and generalization ability to demonstrate the superiority of the proposed method.
\end{abstract}


\section{Introduction}

Robotic grasping or grasp detection \cite{newbury2022deep,doi:10.1146/annurev-control-062122-025215} aims 
to detect graspable points of a scene and recognizes 
their corresponding parameter configurations of a gripper that can induce successful grasping behaviors.
Grasp detection is a fundamental skill for robot manipulation tasks, which has
such broad applications as clutter removal \cite{9561073}, 
transfer of surgical instruments \cite{7015593}, intelligent logistics \cite{song2021robot}, manufacturing \cite{oliver2020robotic},
service robots \cite{9541299}, and so on.

State-of-the-art methods of robotic grasping formulate it as a machine learning problem \cite{NEURIPS2020_994d1cad}, 
and use an end-to-end trained deep neural network to predict graspable points and their corresponding 
gripper parameters from visual observations \cite{breyer2021volumetric,9830843}. 
In general, the backbone models are selected according to the input data types.
For instance, the PointNet \cite{Qi_2017_CVPR} and PointNet++ \cite{NIPS2017_d8bf84be} models are commonly used for unstructured point cloud input 
\cite{Fang_2020_CVPR,NEURIPS2020_994d1cad}, and
3D convolutional neural networks (3D-CNN) \cite{Kopuklu_2019_ICCV} is commonly used for structured voxelized space input \cite{breyer2021volumetric}.

However, existing backbone models of robotic grasping that employ convolution operation 
usually learn features in a local region, and require aggregating local features 
layer by layer to obtain a global representation. This process is slow and not efficient
for global feature learning. Considering the importance of global information for grasping an object \cite{9810182},
more efficient backbone models are needed to investigate. Transformer is a recently proposed model \cite{NIPS2017_3f5ee243} 
that leverages attention mechanism to learn both global and local information effectively. 
A transformer model takes a sequence of tokens as input, and process them through 
multiple layers of multi-head attention-based blocks.
In each layer, a local hidden state is represented by summarizing all hidden states weighted by attention scores, 
which can capture global information of the input sequence effectively.
Initially proposed in natural language processing \cite{NIPS2017_3f5ee243,devlin-etal-2019-bert}, 
transformer-based models have nowadays been successfully used in 
many other domains, such as computer vision \cite{dosovitskiy2021an}, decision making \cite{NEURIPS2021_7f489f64}, 
and 3D point cloud processing \cite{Lai_2022_CVPR}. 

Recently, Kan \textit{et al.} \cite{9810182} propose to use a transformer model for 2D grasp detection,
However, for flexible manipulation, 6-Degree-of-Freedom (6-DoF) grasping \cite{9561877} that enables a robot arm to 
pick up objects from arbitrary orientations in a 3D space \cite{murali20206} is needed,
which is much more general than 2D grasping \cite{mahler2017dex,song2020grasping}.
Extending transformer models to 6-DoF grasping in a 3D space is more challenging than 2D grasping for at least three reasons:
1) Visual observation of a 2D scene can be easily represented as a 2D image, \textit{i.e.} a 2D numerical matrix,
while representing a 3D scene has more options, such as point cloud, mesh, voxel, implicit surface, distance function, and so on.
2) Transformer models usually take a 1D sequence as input. How to serialize a 3D representation to a sequence 
while maintaining the 3D information is a non-trivial problem. 3) The multi-head attention 
calculation in a transformer model usually results in heavy computational and memory costs, which is even worse 
for 3D robotic grasping due to the curse of dimension. How to balance feature learning and efficiency of a transformer 
model is challenging.  

In this paper, we propose a transformer-based 6-DoF grasp detection method. 
The general framework follows volumetric grasping network (VGN) \cite{breyer2021volumetric},
where we take a Truncated Signed Distance Function (TSDF) as input, and learn per-voxel feature vectors through 
a transformer model. The per-voxel feature vectors are then used to 
predict grasping parameters with three heads: grasping quality, 
gripper orientation, and gripper width. Instead of using a 3D convolutional neural network as the backbone
model, we propose to use a transformer-based model to learn both global and local feature more efficiently for robotic grasping.
In particular, motivated by \cite{9706678},
the transformer model first serializes the TSDF input as a sequence of feature vectors 
and encodes the sequence with a deep vision
transformer model ViT \cite{dosovitskiy2021an}, which can obtain a set of aggregated hidden feature vectors through multi-head attention.
We then decode the hidden features to obtain per-voxel feature vectors through deconvolution and skip-connections. 
To demonstrate the effectiveness of the transformer-based model for 6-DoF grasp detection, we conducted a set of experiments
on a recently proposed pile and packed dataset \cite{jiang2021synergies}. Experiment results show that 
our transformer-based model can achieve higher success rates and declutter rates.  
The contributions of this paper are:

\begin{enumerate}
    \item We identify the problem of machine learning based robotic grasping, and propose to use transformer-based models for 
    6-DoF grasp detection.  
    \item Motivated by existing methods \cite{9706678,breyer2021volumetric}, we propose a transformer-based robotic grasping method,
    which can learn both global and local feature effectively in a 3D space.
    \item On a recently introduced pile and packed dataset, we demonstrate that our method outperforms existing 6-DoF robotic grasping
    methods significantly, which indicates the potential of using transformer models for 6-DoF grasping.
\end{enumerate}

\section{Related Work}



For state-of-the-art performance, in this section, 
we only consider deep learning-based methods and 
give a brief review of 6-DoF robotic grasping according to the input types: unstructured
point cloud and structured voxelized space. For more thorough reviews of robotic grasping, 
readers can refer to \cite{newbury2022deep,doi:10.1146/annurev-control-062122-025215,2202.03631} and the references therein.

\subsection{6-DoF Grasping on Point Cloud}

A point cloud is a set of point coordinates, and each point is possibly associated 
with some properties, such as RGB colors. Point cloud is an irregular geometric data format, 
and learning its feature requires permutation invariance of points and transformation invariance of 
the whole cloud \cite{Qi_2017_CVPR}. PointNet \cite{Qi_2017_CVPR} is the first deep learning model to 
learn both global and per-point local feature embeddings of a point cloud through multi-layer perceptron
and max pooling. PointNet++ \cite{NIPS2017_d8bf84be} extends PointNet with multiscale learning ability through sampling 
and grouping. PointNet \cite{Qi_2017_CVPR} and PointNet++ \cite{NIPS2017_d8bf84be} are commonly used as backbone models 
for 6-DoF robotic grasping.
In \cite{Fang_2020_CVPR}, Fang \textit{et al.} use PointNet to 
encode an input point cloud to obtain the vector representation of each point, and then 
learn the approaching vectors, operation parameters, and tolerance scores for each point.
Subsequentially, the executed grasping points are sampled and filtered by objectiveness and tolerance.
Similarly, in \cite{NEURIPS2020_994d1cad}, Wu \textit{et al.} propose Grasp Proposal Networks (GPNet), which uses 
PointNet++ to encode an input point cloud. Different from \cite{Fang_2020_CVPR}, GPNet parameterizes a grasp as 
the two contact points of the gripper, and generates a set of neighbor points for each point of the cloud 
to predict its corresponding peer contact point. 
Due to the unstructured characteristics and the large volume of a point cloud, 
point cloud-based methods usually need to sample a subset of points as grasp candidates, 
which results in performance drop. In \cite{9711469},
Wang \textit{et al.} propose to learn \textit{graspness} property per-point, and use it to filter out non-graspable points,
which is shown to improve grasping success rates significantly on the GraspNet-1Billion dataset. 
In \cite{9830843}, Alliegro \textit{et al.} adopt a differentiable sampler and train it end-to-end with grasp regression, which is shown 
to improve both the performance and running time compared to \cite{NEURIPS2020_994d1cad}. 

PointNet mainly relies on local feature learning to aggregate global information progressively, it is still not efficient
for robotic grasping, which requires an effective encoding of the global information of an input. 
In computer vision and graphics, researchers
have explored the use of transformer models on point cloud processing \cite{2205.07417}, such as point cloud segmentation \cite{Lai_2022_CVPR},
classification \cite{9880161}, and shape completion \cite{Yu_2021_ICCV}.
However, the number of points in a point cloud input is not fixed and too high to be processed with multi-head attention efficiently,
which is especially serious in a real robot scenario. 
How to use transformer models in robotic grasping is still a challenging problem.

\subsection{6-DoF Grasping on Voxelized Space}

Compared to point cloud, voxelized representation partitions a 3D space into uniform grids, which is a structured 
geometric data format. 
Commonly used voxelized representation include 
voxelized occupancy grid \cite{8206060,8968115} and 
voxelized Truncated Signed Distance Function (TSDF) \cite{breyer2021volumetric,jiang2021synergies}.
In \cite{8206060}, Varley \textit{et al.} use a voxelized occupancy grid to represent objects
for shape completion. The authors train a 3D-CNN model and during testing, the model takes an occluded point cloud input 
and completes the shape for grasp planning. Similarly, in \cite{8968115}, Liu \textit{et al.} consider a voxelized occupancy grid
input and use 3D-CNN to predict graspable points. They further introduce 
a consistency loss to ensure a one-to-one mapping from objects to grasp poses. Another option for voxelized 3D space
representation is voxelized Truncated Signed Distance Function (TSDF), where each grid denotes a truncated signed distance
to the object surface. In \cite{breyer2021volumetric}, Breyer \textit{et al.} propose Volumetric Grasping Network (VGN) that 
takes a TSDF as input, and use a 3D-CNN
model to learn per-voxel feature vectors, which are then used to regress the grasping quality, gripper orientation,
and gripper width for each voxel. 
However, either voxelized occupancy grid or TSDF require multi-views to integrate a 3D representation.
Jiang \textit{et al.} extend VGN by introducing an auxiliary shape completion task, which 
only requires one view to make grasp predictions. The authors also report performance improvement because of the use of 
multi-task learning.

Existing methods of 6-DoF grasping on voxelized space intensively employ 3D-CNN as backbone models for grasp detection 
\cite{doi:10.1146/annurev-control-062122-025215}. However, convolution operation also suffers from the 
local learning problem, which is not efficient to aggregate global feature for robotic grasping.
Because voxelized space can be viewed as 3D images, it is relatively easy to use transformer models to replace 3D-CNN
as the backbone model, which is the focus of this paper.

\section{Method}

\begin{figure}
    \centering
    \includegraphics[width=1.0\linewidth]{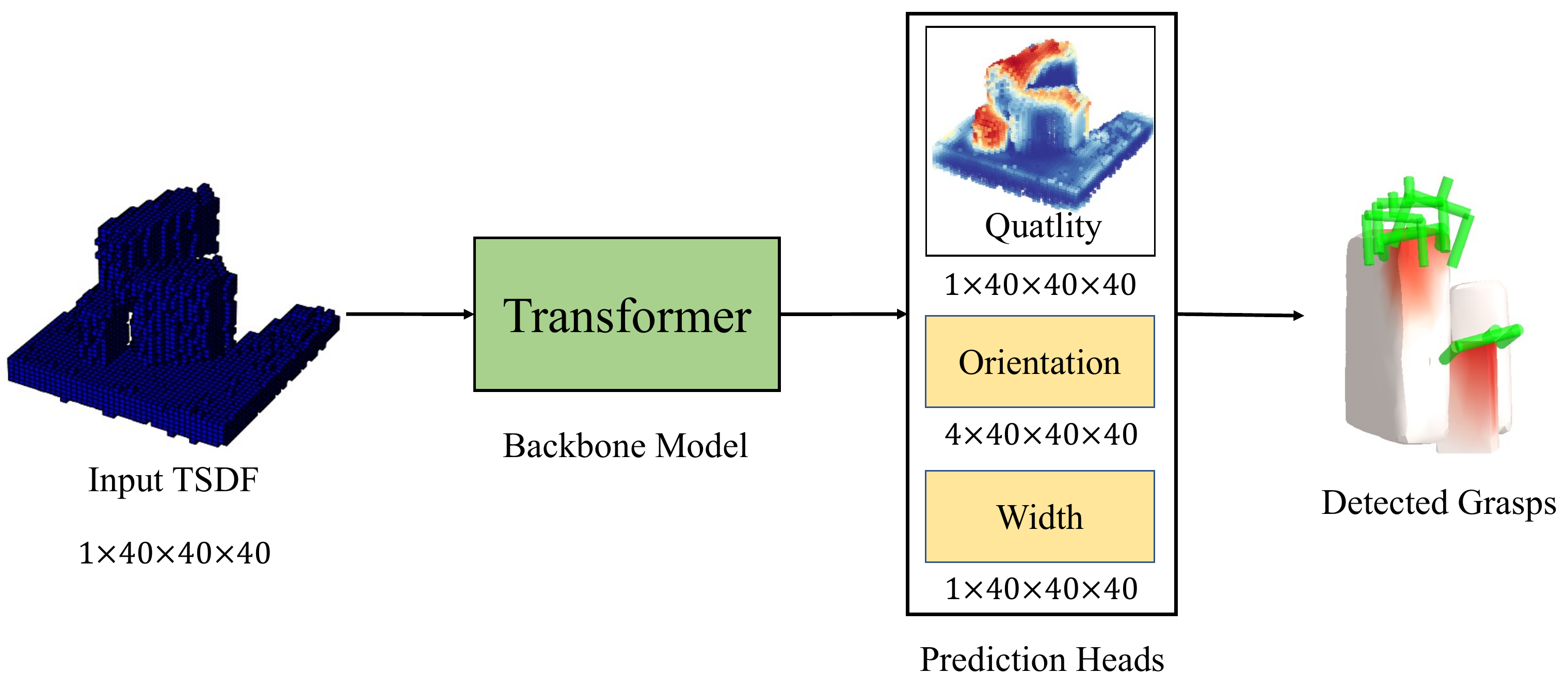}
    \caption{Overiew of our transformer-based 6-DoF grasp detection method.}
    \label{overview}
\end{figure}

The overview framework of our method is shown in Figure \ref{overview} 
\footnote{For the convenience of reference, in Figure \ref{overview} and \ref{unetr}, we annotate the 
exact tensor dimensions used in the implementation.}.
Similar to volumetric grasping network (VGN) \cite{breyer2021volumetric}, the input is a Truncated Signed Distance Function (TSDF) representation ${\bf x}\in \mathbb{R}^{1\times N\times N \times N}$,
where $N$ denotes the resolution of voxels and ${\bf x}(i,j,k) = {\bf x}(0, i, j, k) \in [0, 1]$ denotes the distance value indexed by 
$i,j,k \in \{0, 1, \ldots, N-1\}$.
Through a transformer-based backbone network, we learn per-voxel feature vectors as 
${\bf y} \in \mathbb{R}^{D\times N\times N \times N}$,
where ${\bf y}(i,j,k) \in \mathbb{R}^D$ and $D$ denotes the dimension of per-voxel feature vector. 
We then use three prediction heads to fit the 
grasping parameters, namely, grasping quality, gripper orientation, and gripper width, similar to VGN \cite{breyer2021volumetric}. 
In the following sections, we introduce the transformer-based 
backbone model, prediction heads, and training and grasp detection in details.

\subsection{Backbone Network}

\begin{figure*}[!t]
    \centering
    \includegraphics[width=1.0\linewidth]{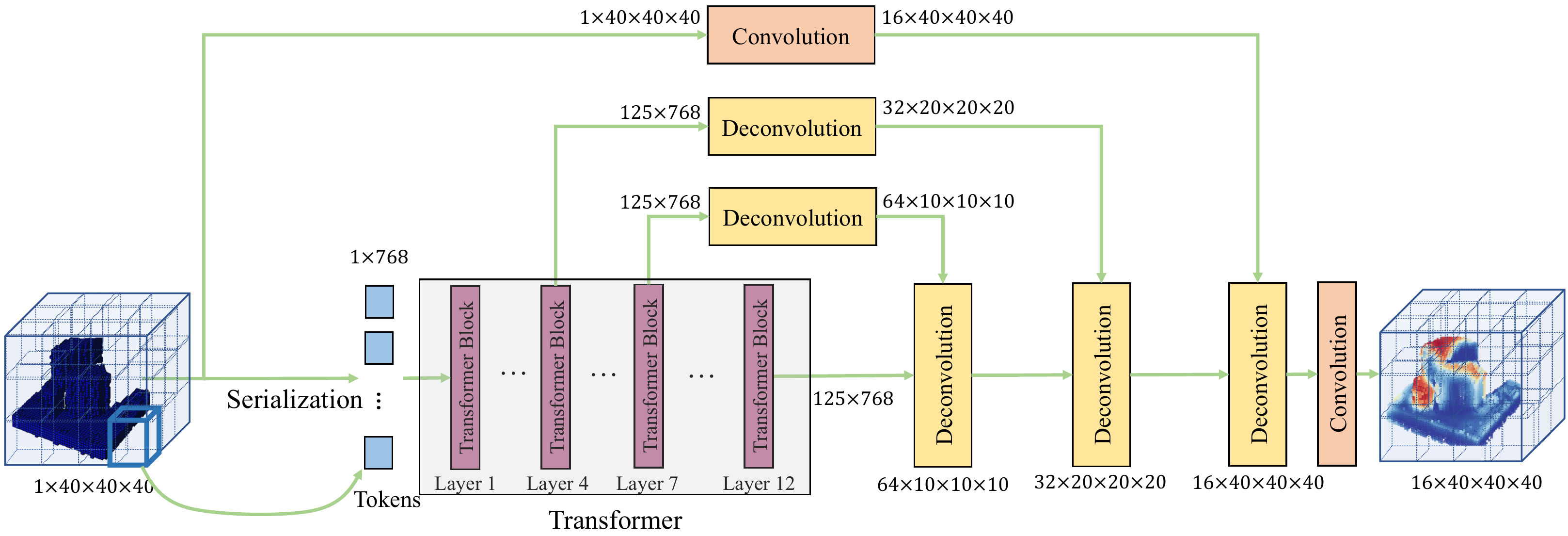}
    \caption{The architecture of the transformer model for encoding a TSDF input.}
    \label{unetr}
\end{figure*}

We adopt UNETR \cite{9706678} as the backbone network, which is a transformer-based model initially used for 
medical image segmentation. We tune the architecture of UNETR to fit our robotic grasping problem.
As shown in Figure \ref{unetr}, the backbone network is a UNet architecture \cite{9053405} that first 
serializes an input TSDF, and then encodes the sequence with a vision transformer ViT \cite{dosovitskiy2021an} 
to obtain a global representation. 
We then decode the global representation to obtain per-voxel representations via deconvolution and skip-connections. 

\subsubsection{Serialization}

Transformer models require a sequence of tokens as input. We need to first serialize the TSDF input as a sequence.
Moreover, although the attention mechanism can efficiently aggregate global information, 
it results in heavy computational and memory costs, which is square complexity in terms of the input sequence length. Therefore,
we need to limit the cardinality of the input sequence. Similar to \cite{9706678}, as shown in Figure \ref{unetr},
we treat each non-overlapping cubic patch with size $C$ of the input TSDF as one token, 
and linearly project each token as a 1D vector. In total, we can obtain $M=(\frac{N}{C})^3$ tokens, and adjust the number of 
tokens by choosing different $C$ values.
We then serialize all tokens as
${\bf x}' = (x'_0, x'_1, \ldots, x'_{M-1}) \in \mathbb{R}^{D' \times M}$, where $x'_i \in \mathbb{R}^{D'}$,
$D'$ is the dimension of the projected embedding. ${\bf x}'$ is then transformed with a 
weight matrix $W$ and we add a learnable position embedding $P$ as:

\begin{equation}
    {\bf z}^{(0)} = (Wx'_0, Wx'_1, \ldots, Wx'_{M-1}) + P,
\end{equation}

\noindent where $W\in \mathbb{R}^{K\times D'}$ and $P \in \mathbb{R}^{K\times M}$. 
${\bf z}^{(0)}$ is the input token sequence of the transformer model.

\subsubsection{Encoding with Transformer Blocks}

The transformer model consists of $L$ transformer blocks and each transformer block transforms
input state ${\bf z}^{(i)}=(z^{(i)}_0, z^{(i)}_1, \ldots, z^{(i)}_{M-1})$ to output state 
${\bf z}^{(i+1)}=(z^{(i+1)}_0, z^{(i+1)}_1, \ldots, z^{(i+1)}_{M-1})$ through a multi-head self-attention
block and a feedforward block, where $z^{(i)}_j, z^{(i+1)}_j\in \mathbb{R}^{K}$.

\paragraph{Multi-head Self-Attention} 

Given an input sequence ${\bf z}^{(i)}=(z^{(i)}_0, z^{(i)}_1, \ldots, z^{(i)}_{M-1})$, Multi-head Self-Attention (MSA) 
processes it as:

\begin{equation}
    {\bf z}^{(i+1)} \leftarrow \text{MSA}(\text{LN}({\bf z}^{(i)})) + {\bf z}^{(i)},
\end{equation}

\noindent where $\leftarrow$ denotes value assignment, MSA denotes a MSA operation, and LN denotes a layernorm operation.
For each hidden state in $z^{(i)}_j \in {\bf z}^{(i)}$, where $j\in \{0, 1, \ldots, M-1\}$,
MSA calculates it as:

\begin{equation}
    \begin{aligned}
        z^{(i+1)}_j \leftarrow & \biggl( \text{A}_{W_0^k,W_0^q,W_0^v}({\bf z}^{(i)}, z^{(i)}_j), \ldots, \\
        &  \text{A}_{W_{H-1}^{k},W_{H-1}^{q},W_{H-1}^{v}}({\bf z}^{(i)}, z^{(i)}_j) \biggr) W_o,
    \end{aligned}
\end{equation}

\noindent where $W_h^k,W_h^q,W_h^v,W^o$ are the parameters of MHA,
$H$ is the number of heads, $h \in \{0,1,\ldots,H-1\}$, 
$W_h^k,W_h^q,W_h^v \in \mathbb{R}^{\frac{K}{H} \times K}$, $W^o 
\in \mathbb{R}^{K \times K}$, and 
$\text{A}_{W_h^k,W_h^q,W_h^v}({\bf z}^{(i)}, z^{(i)}_j)$ is one head transformation,
which is calculated as:

\begin{equation}
    \text{A}_{W_h^k,W_h^q,W_h^v}({\bf z}^{(i)}, z^{(i)}_j)= 
    \sum_{m=0}^{M-1} \alpha_m W_h^v z^{(i)}_m,
\end{equation}

\noindent where $\alpha_m = \text{softmax}(\frac{({W_h^q z^{(i)}_j})^T W_h^k z^{(i)}_m}{\sqrt{K/H}})$.



\paragraph{Feedforward Block} 

The obtained ${\bf z}^{(i+1)}$ from multi-head self-attention is then processed with a feedforward
block as follows:

\begin{equation}
    {\bf z}^{(i+1)} \leftarrow \text{MLP}(\text{LN}({\bf z}^{(i+1)})) + {\bf z}^{(i+1)},
\end{equation}

\noindent where MLP denotes a multi-layer perception operation, and LN denotes a layernorm operation.



\subsubsection{Decoding with Deconvolution and Skip-Connections}

Through multiple layers of transformer blocks, we can obtain hidden representations with different layers, 
and each layer represents aggregated information at a specific scale. Considering $L$ layers of transformer blocks,
we obtain the final layer hidden state as ${\bf z}^{L-1} \in \mathbb{R}^{K\times M}$. 
We project it as a 3D voxelized patch 
${\bf y}^{(0)} = \text{Proj}({\bf z}^{L-1}) \in \mathbb{R}^{M \times \sqrt[3]{K} \times \sqrt[3]{K} \times \sqrt[3]{K} }$,
where $\text{Proj}$ denotes the projection operation.

The serialization module projects non-overlapping patches as tokens, which can be viewed as down-sampling the initial TSDF.
Therefore, to obtain per-voxel feature vectors of the input TSDF,
we up-sample ${\bf y}^{(0)}$ with deconvolution \cite{Noh_2015_ICCV}. Moreover, to merge multiscale hidden states,
similar to UNet \cite{9053405}, we select several layers of the transformer blocks, project them as 3D voxelized patches,
up-sample them with deconvolution, and skip-connect them with the decoded state layer by layer as follows:

\begin{equation}
    {\bf y}^{(i+1)} \leftarrow \text{DECONV} \biggl( \text{DECONV}({\bf y}^{(i)}) \oplus  \text{DECONV}(\text{Proj}({\bf z}^{l})) \biggr),
\end{equation}

\noindent where $\text{DECONV}$ denotes deconvolution operation and $l \in \{0, 1, \ldots, L-1\}$ is the selected layer.
One exception is the final decoding layer ${\bf y}^{(L'-1)}$, where we concatenate with the input TSDF processes by convolution
and obtain the final output ${\bf y}$ as:

\begin{equation}
    {\bf y} = \text{CONV} \biggl( {\bf y}^{(L'-1)} \oplus \text{CONV}({\bf x}) \biggr),
\end{equation}

\noindent where $\text{CONV}$ denotes convolution operation, $L'$ is the number of decoding layer, and
${\bf y} \in \mathbb{R}^{D\times N\times N \times N}$ is the feature representation of the input TSDF.


\subsection{Grasping Heads}

For each voxel, we predict whether it can be used as a graspable position and its corresponding grasping parameters.
Similar to VGN \cite{breyer2021volumetric}, we add quality ${\bf Q} \in [0,1]^{1\times N \times N \times N}$, 
gripper orientation ${\bf R} \in \mathbb{R}^{4\times N \times N \times N}$, 
and gripper width ${\bf W} \in \mathbb{R}^{1\times N \times N \times N}$ heads through convolution, \textit{i.e.},

\begin{equation}
    \begin{aligned}
        {\bf Q} &= \text{CONV}_1 ({\bf y}) \\
        {\bf R} &= \text{CONV}_2 ({\bf y}) \\
        {\bf W} &= \text{CONV}_3 ({\bf y}) 
    \end{aligned},
\end{equation}

\noindent For each voxel, quality $q$ is a score indicating the quality of grasping. 
High quality score indicates high probability of successful grasping. 
Orientation ${\bf r}$ denotes the orientation of a gripper at a specific voxel to execute grasping, which 
is parameterized with quaternion, \textit{i.e.} ${\bf r} \in \mathbb{R}^4$. 
Gripper width $w_i$ denotes the gripper width before executing grasping.

\subsection{Training and Grasp Detection}

\subsubsection{Training}

For each voxel, 
considering a groundtruth grasp $g_i = (q_i, {\bf r}_i, w_i)$ 
and the predicted grasp on this voxel $\hat{g}_i = (\hat{q}_i, {\bf \hat{r}}_i, \hat{w}_i)$,
the training objective is calculated as follows \cite{breyer2021volumetric,jiang2021synergies}:

\begin{equation}
    \mathcal{L}(\hat{g}_i, {g}_i) = \mathcal{L}_q({q}_i, q_i) + q_i(\mathcal{L}_r(\hat{r}_i, {r}_i)+\mathcal{L}_w(\hat{w}_i, {w}_i)),
\end{equation}

\noindent where $q_i \in \{0, 1\}$ is the groundtruth grasping label, 
${\bf r}_i$ is the groundtruth rotation quaternion, $w_i$ is the groundtruth gripper width,
$\mathcal{L}_q$ denotes the binary cross-entropy loss, $\mathcal{L}_r = \min (\mathcal{L}_{quat}({\bf \hat{r}}_i, {\bf r}_i), \mathcal{L}_{quat}({\bf \hat{r}}_i, {\bf r}_{i \pi}))$,
$\mathcal{L}_{quat}({\bf \hat{r}}, {\bf r})) = 1-|{\bf \hat{r}} \cdot {\bf r}|$,
${\bf r}_{i \pi}$ denotes the rotation quaternion by rotating ${\bf r}_i)$ $180^{\circ}$ along the gripper's wrist axis,
and $\mathcal{L}_w$ denotes the mean square error.

\subsubsection{Grasp Detection}

After obtaining a trained model, we can use the predicted quality tensor to select 
graspable candidates. In particular, for a voxel with quality $q$,
if its value is higher than a predefined threshold $\epsilon$, \textit{i.e.} $q>\epsilon$, 
this voxel is then used to fit gripper rotation and width.
Considering the 
voxel size $v$ and the rigid transformation $T$ between a simulated or real robot platform 
and the TSDF volume, the robot gripper configuration for executing a grasping action can be calculated as:

\begin{equation}
    \begin{aligned}
        p_g &= T\frac{(i,j,k)^T}{v} \\
        {\bf r}_g &= T {\bf r} \\
        w_g &= \frac{w}{v}
    \end{aligned},
\end{equation}

\noindent where $(i,j,k)$ denotes the indices of a graspable voxel, 
$p_g$ denotes the position of the gripper center,
${\bf r}_g$ denotes the orientation of the gripper, and $w_g$ denotes the width of the gripper. 
$p_g, {\bf r}_g, w_g$ are in the base frame.

\section{Experiment}

To verify the effectiveness of the proposed method, we conducted a set of experiments 
following the setting in \cite{breyer2021volumetric,jiang2021synergies}.

\subsection{Pile and Packed Dataset}

Since the authors in \cite{breyer2021volumetric} have not released their training and test dataset, 
for fair comparison, we adopt the dataset used in \cite{jiang2021synergies}, 
which adopts the same data generation procedure as in \cite{breyer2021volumetric}. We term it as pile and packed dataset.


Pile and packed dataset contains two scenarios for the clutter removal task, which requires to generate a set of 
grasps that can move a clutter of objects to a target bin. Each scenario is constructed 
from 303 training objects following the procedure in \cite{breyer2021volumetric}. The pile scenario
is a fixed-size workspace containing objects dropped randomly. The packed scenario is a fixed-size workspace
containing objects placed randomly at their canonical pose. The statistics of pile and packed 
dataset is shown in Table \ref{pileandpacked}. For each scenario, 
we use 90\% of the grasps as training data and the remaining 10\% of as validation data.
For testing, there are 40 objects that do not appear in the 303 training objects, which are used to 
compose simulation environment on the fly. The pile scenario uses all 40 objects, while the packed 
scenario requires objects to have proper sizes, and only uses 16 objects among them. 

\begin{table}[]
    \centering
    \begin{tabular}{|c|c|c|c|}
    \hline
           & \#scenes & \#grasps & \#test objects\\ \hline
    pile   & 83305    & 1564546 &  40 \\ \hline
    packed & 16640    & 638012  & 16 \\ \hline
    \end{tabular}
    \caption{Statistics of pile and packed dataset.}
    \label{pileandpacked}
\end{table}

\subsection{Baselines}

Similar to \cite{jiang2021synergies}, we adopt SHAF, GPN, and VGN as baselines.
SHAF uses heuristic to select objects with the highest point to grasp.
GPN \cite{7759114} is a classical hypothesizing and testing method \cite{doi:10.1146/annurev-control-062122-025215} that first samples 
grasping poses in vicinity of points in an input point cloud, and then classifies whether the sampled grasp works or not.
VGN \cite{breyer2021volumetric} is the main baseline that we target to compare with, which uses a 
3D-CNN as its backbone model to learn per-voxel features. 

\subsection{Metrics}

We adopt grasp success rates (GSR) and declutter rates (DR) as evaluation metrics. 
GSR is calculated as the ratio between the number of successful grasping $n_\text{suc}$
and the number of grasps predicted $n_\text{pred}$: $GSR=\frac{n_\text{suc}}{n_\text{pred}}$. DR is calculated as the ratio 
between the number of removed objects $n_\text{rem}$ and the total number of objects $n_\text{obj}$ in the current scene:
$DR=\frac{n_\text{rem}}{n_\text{obj}}$.

\subsection{Implementation}

We implement our transformer-based 6-DoF grasping model in PyTorch \footnote{\url{https://pytorch.org}}. 
Similar to \cite{breyer2021volumetric}, the input TSDF is divided into $40\times 40\times 40$ grids, \textit{i.e.} $N=40$.
To save GPU memory and computational time, we choose a patch size $C=8$, which results in $M = 5\times 5 \times 5 = 125$ tokens.
Similar to \cite{jiang2021synergies}, we train the two scenarios pile and packed separately, and use the same setting
in the remaining experiments. As shown in Figure \ref{unetr}, we use a transformer model with $L=12$ layers to encode the input TSDF,
and decode it with $L'=3$ deconvolution blocks.
Empirically, we choose the convoluted input TSDF and the outputs of layer 4, 7 
of the transformer model to skip-connect to the three deconvolution blocks in sequence. 
Adam optimizer is used with learning rate 
$0.0001$ and batch size $64$. We set $K=768, H=12, D=16, \epsilon=0.9$.
The model is trained 21 epochs and we use the checkpoint of the last epoch for evaluation.
For training the models,
we use a server with Intel(R) Xeon(R) Silver 4210R CPU @ 2.40GHz, 64GB memory, 
and NVIVIDA GeForce 3090 GPU ($\times 1$). 
The training time lasts about 60 hours for each scenario.

\subsection{Results and Analysis}\label{results}

Similar to \cite{jiang2021synergies}, for testing each scenario, 
we ran 100 simulation rounds, and repeated five times with fixed seeds.
We report the mean and standard deviation of the results. For  SHAF, GPD, we take the results from \cite{jiang2021synergies}.
For VGN, we take the results from \cite{jiang2021synergies} (denoted as VGN) and also evaluate the results in the same setting 
with the pre-trained models released in \cite{jiang2021synergies} (denoted as VGN-R).
The simulation results are shown in Table \ref{simulationresults}.

From the results we can find that our method can significantly outperform SHAF, GPD, VGN on both 
grasp success rates (GSR) and declutter rates (DR). In particular, although our method and VGN share the same framework,
because of the use of a transformer model, our method can aggregate global information of the visual observation more efficiently,
and can achieve about 5\% improvement on both GSR and DR. Similar to \cite{breyer2021volumetric},
we also adjust the quality threshold $\epsilon$ as $0.8$ and $0.95$ of our method, and we find that performance
results are similar to the ones in \ref{simulationresults}, which also indicates that our method can predict graspable
voxels with higher probability.
\begin{table}[]
    \centering
    \begin{tabular}{|c|cc|cc|}
    \hline
    \multirow{2}{*}{Method} & \multicolumn{2}{c|}{Packed}           & \multicolumn{2}{c|}{Pile}             \\ \cline{2-5} 
                            & \multicolumn{1}{c|}{GSR(\%)} & DR(\%) & \multicolumn{1}{c|}{GSR(\%)} & DR(\%) \\ \hline
    SHAF                    & \multicolumn{1}{c|}{$56.6\pm2.0$}    & $58.0\pm3.0$   & \multicolumn{1}{c|}{$50.7\pm1.7$}    & $42.6\pm2.8$   \\ \hline
    GPD                     & \multicolumn{1}{c|}{$35.4\pm1.9$}    & $30.7\pm2.0$   & \multicolumn{1}{c|}{$17.7\pm2.3$}    & $9.2\pm1.3$    \\ \hline
    VGN                     & \multicolumn{1}{c|}{$74.5\pm1.3$}    & $79.2\pm2.3$   & \multicolumn{1}{c|}{$60.7\pm4.2$}    & $44.0\pm4.9$   \\ \hline
    VGN-R                     & \multicolumn{1}{c|}{$74.4\pm3.6$}    & $79.8\pm3.1$   & \multicolumn{1}{c|}{$62.5\pm2.4$}    & $46.4\pm2.9$   \\ \hline
    Ours & \multicolumn{1}{c|}{${\bf 79.7}\pm 2.0$}    & ${\bf 85.9}\pm1.7$   & \multicolumn{1}{c|}{${\bf 66.9}\pm 3.0$}        &  ${\bf 50.9}\pm 3.5$      \\ \hline
    \end{tabular}
    \caption{The comparison of simulation results on packed and pile scenarios.}
    \label{simulationresults}
\end{table}

\subsection{Discussion}

We further evaluate the running time and generalization ability of the proposed method.

\subsubsection{Running Time}

Since robotic grasping is a fundamental task of robot manipulation, it is critical to generate successful
grasps in real-time. With the experiment setting used in section \ref{results}, we calculate the average time of 
generating all grasps for one input TSDF. The results are shown in Table \ref{time}.
Because the computation process of a transformer model is relatively heavy, our method 
nearly doubles the running time of VGN. However, it is still in a reasonable range for real-time usage.
Moreover, since transformers gradually become universal models for machine learning, dedicated 
devices for transformers are available \cite{hanruiwang2020hat} for further improving the performance.

\begin{table}[!t]
    \centering
    \begin{tabular}{|c|c|c|}
    \hline
         & Packed & Pile  \\ \hline
    VGN-R  &  $11.0\pm 0.5$  & $11.2\pm 0.6$  \\ \hline
    Ours &   $24.3\pm 0.7$  & $24.6\pm 1.1$  \\ \hline
    \end{tabular}
    \caption{The comparison of average time for generating all grasps of one input TSDF in milliseconds (ms).}
    \label{time}
\end{table}



\subsubsection{Generalization on New Scenarios}

Another important aspect for real-world usage of robotic grasping is the generalization ability of a model 
on new scenarios. With the experiment setting used in section \ref{results}, we test the performance of
the model on the packed scenario, while the model is trained with the pile data (denoted as Pile$\rightarrow$Packed),
and the performance of
the model on the pile scenario, while the model is trained with the packed data (denoted as Packed$\rightarrow$Pile).
The results are shown in Table \ref{generalization}. 
In general, the performance drop for both VGN and our method.
However, it is interesting to find that our method still outperform VGN significantly, especially for the 
packed scenario, where GSR and DR of our method are higher than the ones of VGN for about 10\%.
This indicates that our method can generalize better than VGN on new scenarios, 
which is more suitable for real usage.   

\begin{table}[]
    \centering
    \begin{tabular}{|c|cc|cc|}
    \hline
    \multirow{2}{*}{Method} & \multicolumn{2}{c|}{Pile$\rightarrow$Packed}           & \multicolumn{2}{c|}{Packed$\rightarrow$Pile}             \\ \cline{2-5} 
                            & \multicolumn{1}{c|}{GSR(\%)} & DR(\%) & \multicolumn{1}{c|}{GSR(\%)} & DR(\%) \\ \hline
    VGN-R                     & \multicolumn{1}{c|}{$60.1\pm1.8$}    & $63.2\pm1.8$   & \multicolumn{1}{c|}{$44.9\pm1.2$}    & $33.5\pm1.3$   \\ \hline
    Ours & \multicolumn{1}{c|}{${\bf 74.0}\pm 2.8$}    & ${\bf 78.9.1}\pm2.7$   & \multicolumn{1}{c|}{${\bf 49.5}\pm 1.4$}        &  ${\bf 39.8}\pm 1.7$      \\ \hline
    \end{tabular}
    \caption{The comparison of generalization performance on new scenarios.}
    \label{generalization}
\end{table}

\section{Conclusion}

This paper considers 6-DoF grasp detection problem, and we propose to use 
a transformer-based model to learn per-voxel feature to regress grasping parameters.
Through an intensive experiments, we demonstrate that transformer models are good alternatives for 
6-DoF robotic grasping, similar to the case in 2D \cite{9810182}. However, due to the curve of dimension, 
transformer models are not efficient for processing high resolution voxel input. It is also 
interesting to investigate how to utilize transformer models to process unstructured point cloud data 
for robotic grasping tasks. We leave them as future works.

\bibliographystyle{IEEEtran}
\bibliography{ref}

\end{document}